# A Data-Centric Framework for Intraoperative Fluorescence Lifetime Imaging for Glioma Surgical Guidance

Silvia Noble Anbunesan, Mohamed Abul Hassan, *Member, IEEE*, Jinyi Qi, *Fellow, IEEE*, Lisanne Kraft, Han Sung Lee, Orin Bloch, and Laura Marcu, *Senior Member, IEEE*

***Abstract*— Accurate intraoperative assessment of glioma infiltration is essential for maximizing tumor resection while preserving functional brain tissue. Fluorescence lifetime imaging (FLIm) offers real-time, label-free biochemical contrast, but its clinical utility is challenged by biological heterogeneity, class imbalance, and variability in histopathological labeling. We present a data-centric AI (DC-AI) framework that integrates confident learning (CL), class refinement, and targeted label evaluation to develop a robust multi-class FLIm classifier for glioblastoma (GBM) resection margins. FLIm data were collected from 192 tissue margins across 31 newly diagnosed IDH-wildtype GBM patients and initially labeled into seven tumor cellularity classes by an expert neuropathologist. CL was applied to quantify FLIm point-level confidence, identify label inconsistencies, and guide iterative class merging into a three-class scheme ('low', 'moderate', 'high'). The resulting high-fidelity dataset enabled training a model that achieved 96% accuracy in the three-class task. SHAP analysis revealed class-specific FLIm feature importance highlighting distinct optical signatures across the infiltration spectrum. Targeted FLIm analysis further identified biological (e.g., gray matter composition) and acquisition-related (e.g., blood contamination) contributors to low-confidence predictions. Blinded re-evaluation of margins flagged by CL demonstrated intra-pathologist variability, underscoring the value of selective relabeling rather than exhaustive review. Together, these findings demonstrate that a DC-AI framework can systematically improve data reliability, enhance model robustness, and refine biological interpretation of FLIm signals, supporting the development of clinically actionable optical tools for real-time glioma margin assessment.**

## I. Introduction

A main surgical challenge in the resection of diffuse gliomas is the accurate intraoperative delineation of tumor infiltration at the resection margins, where neoplastic cells gradually blend into the surrounding brain parenchyma. The neurosurgeon's goal is to maximize the extent of resection while preserving eloquent brain regions [1]. This requires real-time intraoperative feedback on tumor presence at resection edges to guide the neurosurgeon in deciding whether to preserve or excise tissue.

In recent years, several optical and imaging-based technologies have been developed to address this clinical need. Among them, fluorescence lifetime imaging (FLIm) has emerged as a powerful, label-free spectroscopic technique that enables real-time, in vivo tissue characterization based on intrinsic biochemical and metabolic properties [2]. By capturing differences in tissue autofluorescence [3], [4], FLIm supports the neurosurgeon's goal of achieving maximal safe resection. A FLIm-based binary classification model has demonstrated high accuracy (>87%) in distinguishing tumor from non-tumor regions during glioma surgery, highlighting its potential for intraoperative guidance [5].

However, glioma infiltration is more accurately represented as a continuous spectrum of tumor cellular density rather than a binary classification. FLIm may be sensitive to this biochemical heterogeneity, which would be useful information for the neurosurgeon to have especially while resecting tumor close to eloquent brain regions. A FLIm-based multiclass framework facilitates targeted model improvements that may ultimately improve binary classifiers.

Developing a robust and interpretable multi-class FLIm-based classification framework for intraoperative glioma assessment poses several challenges:

**1. Real-World Clinical Data and Sources of Variability**

The collection of optical data during neurosurgical procedure is subject to multiple sources of variability that can influence FLIm signals beyond just the presence of tumor cells. These sources include biological factors such as tissue type (white or grey matter), necrosis, and local pH, all of which can alter fluorescence lifetimes [3], [4]. Intraoperative artifacts, including blood in the field-of-view and tissue motion, can further introduce noise into the FLIm measurements [6], [7]. In addition, the data is inherently imbalanced, with an overrepresentation of margins exhibiting low tumor cellular density. Such imbalance can lead to biased model performance and reduced sensitivity to clinically important but underrepresented high tumor cellular density cases.

This study was supported by the National Institutes of Health (R01CA250512, P41EB032840), the University of California, Davis, and the Comprehensive Cancer Center's Brain Malignancies Innovation Group funded through the Comprehensive Cancer Center Support Grant (P30CA093373). All authors are affiliated with the University of California, Davis. Silvia Noble Anbunesan, Mohamed Abul Hassan, Jinyi Qi, and Lisanne Kraft, work within the Department of Biomedical Engineering. Han Sung Lee works within the Department of Pathology and Laboratory Medicine. Orin Bloch practices oncologic surgery within the Department of Neurological Surgery. Laura Marcu works within the Department of Neurological Surgery and the Department of Biomedical Engineering. Corresponding Author Laura Marcu (e-mail: lmarcu@ucdavis.edu).

**2. Margin-Level Labels vs. Point-Level Data**

Tumor cellular density labels were assigned at the margin level based on histopathological assessment, whereas FLIm data were acquired at a much finer spatial resolution, capturing fluorescence signals at the individual point level. Consequently, all FLIm points within a given margin inherit the same tumor cellular density label despite potential biochemical and cellular heterogeneity within the margin itself. This mismatch between the high spatial sensitivity of FLIm and the coarser resolution of histological labeling introduces systematic label noise at the point level and complicates model training and validation. To mitigate this, aggregation methods such as majority voting were employed to reconcile predictions across these differing spatial resolutions.

**3. Uncertainty in Histopathological Ground Truth Labels**

Although histopathology is considered the gold standard for tumor assessment, its reliability may be compromised by subjectivity in pathologist interpretation, limited biopsy sampling [8], and tumor heterogeneity [9]. Consequently, some margin-level labels may be inaccurate or ambiguous, leading to mislabeling of all associated FLIm point measurements. This uncertainty introduces additional noise into the training data, posing significant challenges for the development of robust classifiers and the accurate validation of model performance.

Given these challenges, a data-centric (DC) AI approach is essential for developing a robust and biologically meaningful multi-class FLIm classification model. Rather than focusing solely on model architecture, this approach emphasizes data quality [10]. By addressing issues such as class imbalance, label noise, and measurement variability, DC-AI strategies enhance model generalizability and robustness, reduce overfitting, and improve the biological relevance of predictions. This is particularly critical in intraoperative settings where misclassification could directly impact surgical outcomes.

DC-AI methods are increasingly used in medical imaging to address label noise and expert variability, especially in histopathology, where inconsistent annotations and ambiguous ground truth can be mitigated through noise-aware training and multi-expert or uncertainty-modeled labeling strategies [11], [12], [13]. However, these prior approaches focus on spatially resolved histological images and do not directly apply to FLIm's point-level optical measurements with coarse, margin-level pathology labels, motivating the development of a tailored DC framework for FLIm tumor cellular density classification.

In this work, we present a DC-AI methodology for developing a multi-class FLIm classification model that maps glioma infiltration based on tumor cellular density. Utilizing confident learning (CL) techniques, we identify and address sources of class noise, enabling the refinement of pathology scoring and improving the reliability of the training dataset. Our approach also highlights intra-pathologist scoring variability across the dataset and investigates the origin of abnormal FLIm features through targeted FLIm data assessment. By curating a high-fidelity dataset and systematically pruning unreliable samples, we demonstrate that a DC approach is critical to enhancing model performance. This study establishes a robust foundation for the development of next-generation FLIm-based classifiers, supporting more accurate intraoperative decision-making in glioma surgery.

## II. Methods

Fig. 1 outlines the overall methodological pipeline. First, a multi-class classifier was trained using FLIm data acquired from the resection margins of glioma patients to predict tumor infiltration levels. Then, a tailored version of the CL framework [14] was implemented to assess and refine data quality, by identifying low-confidence measurements and potential label errors. Finally, through a combination of data characterization, pruning, and relabeling, a high-fidelity dataset was curated and trained to evaluate the sensitivity of FLIm in detecting tumor infiltration across varying tumor cellular densities.

### *A. FLIm Instrumentation and Data Acquisition*

A pulse sampling FLIm instrument (psFLIm) [15] was used to collect data for this study [16]. Briefly, it employed a 355-nm excitation wavelength Nd-YAG microchip laser (STV-02 E-1X 0, TEEM photonics, France) (0.25µJ pulse energy, 600 ps pulse duration, 460 Hz repetition rate). A 3-meter-long multimodal fiber optic probe (0.22 NA, 365 µm core diameter, 3 mm working distance) delivered the excitation wavelength to the tissue of interest in situ in the resection cavity. The tissue autofluorescence was collected by the same probe and was spectrally resolved by three detectors, each consisting of avalanche photodetectors (APD) modules with an integrated transimpedance amplifier and a high-speed digitizer (2.5 GS/s, NI PXIe-5162, National Instruments, TX), with the following spectral bands: 390/40 nm, 470/28 nm, 540/50 nm, which are centered at the emission maxima of collagen, NAD(P)H, and flavins respectively. For the purposes of this study, we excluded FLIm data from 390/40 nm spectral bands since brain does not have collagen in significant amounts, resulting in poor signals collected.

During tumor resection, the neurosurgeon intraoperatively identified multiple regions (3-10 per patient) within the resection cavity that were suspected to harbor tumor infiltration. These regions, typically measuring 1-2 mm², were scanned using the FLIm system (~300 FLIm points per scan; ~0.8 mm point size). Immediately following imaging, tissue biopsies were obtained from the same sites for histopathological evaluation. The collected specimens were fixed in formalin, embedded in paraffin, sectioned, and stained with hematoxylin and eosin (H&E) using standard histological procedures.

### *B. Study Cohort and Histopathological Labeling*

A total of 96 diffuse glioma patients undergoing surgical resection at the University of California Davis Medical Center were enrolled under an institutional review board (IRB)-approved protocol for the collection of FLIm measurements at the resection margins. A subset of 31 newly diagnosed IDH-wildtype GBM patients (and 192 distinct tissue margins) were chosen as a consistent tumor cohort. Recurrent and lower-grade or IDH-mutant gliomas were excluded to avoid FLIm signal variations associated with differing tumor biology and metabolic profiles [17], [18].

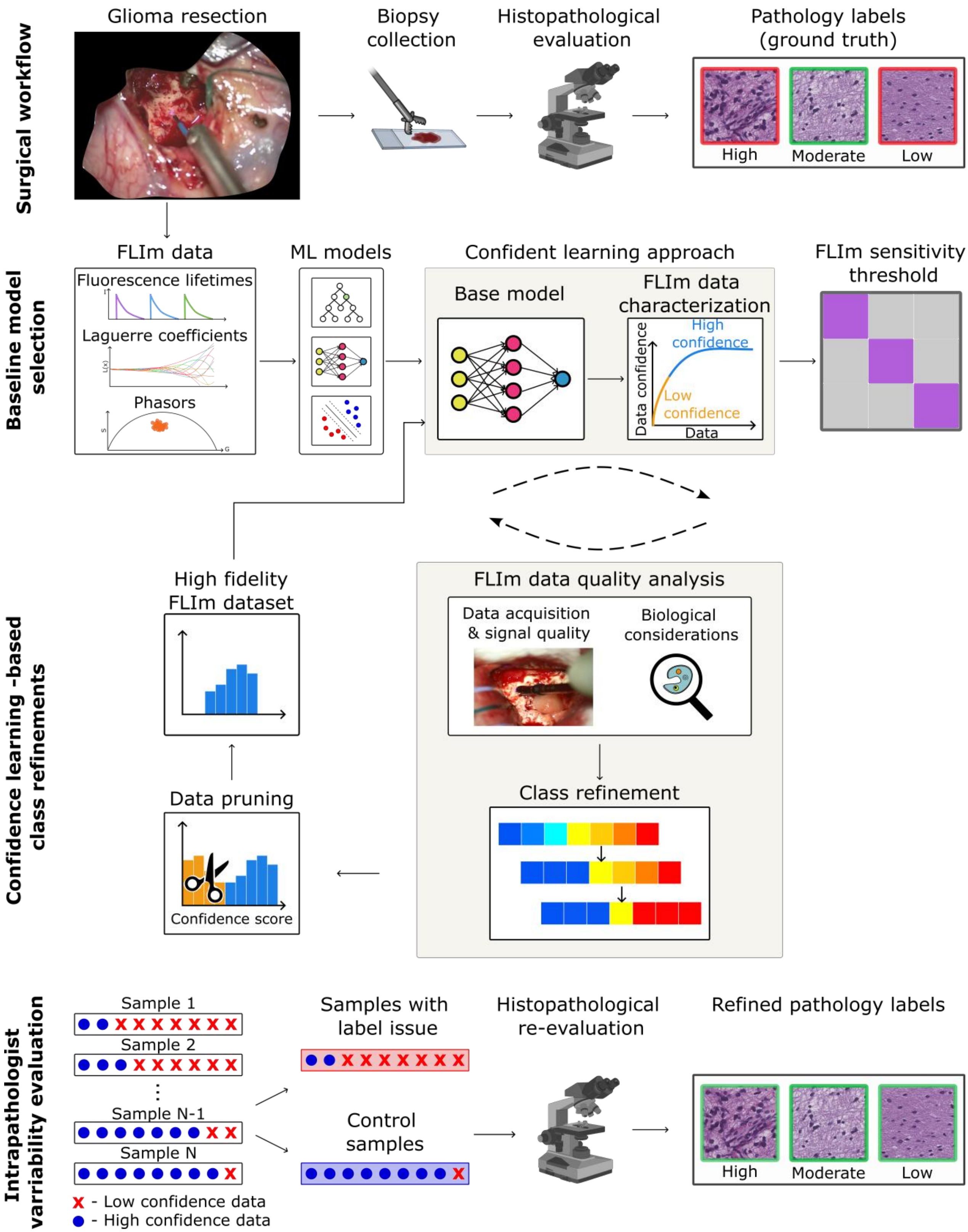

Fig. 1: Overview of the DC-AI methodology for intraoperative fluorescence lifetime imaging (FLIm) in glioma surgery. During tumor resection, several ‘margins’ at the infiltrative edges were scanned using FLIm, and corresponding biopsies were labeled by a neuropathologist based on tumor cellular density. A baseline machine learning model was trained on extracted FLIm features, followed by application of the confident learning algorithm to characterize FLIm data using model posterior probabilities. Low-confidence FLIm data were analyzed to guide iterative class refinements, enabling identification of the FLIm sensitivity threshold. Finally, intra-pathologist variability was assessed by blinded re-evaluation of histopathological labels for margins flagged with potential label issues.

TABLE 1
HISTOPATHOLOGICAL SCORING OF TUMOR CELLULAR DENSITY CLASS

| Tumor Cellular Density Class | Number of Tumor Cells per HPF [a] |
|---|---|
| Absent | 0 |
| Very low | 20-50 present in at least half of the tissue section |
| Low | 50-150 (<10% of surface area) |
| Low-moderate | 150-250 (10-15% of surface area) |
| Moderate | 250-500 (15-25% of surface area) |
| Moderate-high | 500-1000 (25-40% of surface area) |
| High | >1000 (>40% of surface area) |

[a] HPF - high-power field, defined as a 0.5 mm field diameter using a 40x objective.

Histopathological evaluation was conducted by an experienced neuropathologist (H.S.L.), who assessed tumor cellular density in each biopsy specimen, referred to as a "margin", based on tumor cell coverage, morphological atypia, and, when applicable, the Ki-67 proliferation index from H&E-stained sections. Each margin was categorized into one of seven tumor cellular density classes (see Table 1). These labels served as ground truth for training and evaluating the machine learning models. Table 2 summarizes the distribution of patients, margins, and FLIm points across these classes.

### C. FLIm Data Processing and Feature Extraction

TABLE 2
DATASET DISTRIBUTION ACROSS TUMOR CELLULAR DENSITY CLASSES

| Tumor Cellular Density Class | Number of Patients (P) | Number of Margins (N) | Number of FLIm points (n) |
|---|---|---|---|
| Absent | 20 | 43 | 25434 |
| Very low | 25 | 52 | 29733 |
| Low | 23 | 52 | 28088 |
| Low-moderate | 7 | 8 | 4556 |
| Moderate | 12 | 16 | 9307 |
| Moderate-high | 5 | 6 | 3512 |
| High | 10 | 15 | 8097 |
| Total | 31 | 192 | 108727 |

Fluorescence decays were deconvolved using a constrained least-squares deconvolution with Laguerre expansion to extract fluorescence lifetime metrics, including mean lifetimes (LT) and Laguerre coefficients (LC) [19]. Additionally, three phasor harmonics (Ph) per spectral band were computed using fast Fourier transforms of the pulse waveforms [20]. A total of 38 FLIm features from the two spectral bands ($SB_{390}$ and $SB_{470}$) were used to form multidimensional input vectors for classification.

### D. Confident Learning Approach applied to FLIm Data

Given a real-world noisy clinical dataset $D = \{(X_i, Y_i)_{i=1}^{N}\}$ where $X_i = \left\{(x_{i,j})_{j=1}^{n_i}\right\}$ represents the set of FLIm feature vectors acquired from margin $i$ and $Y_i \in \{1, 2, \dots, C\}$ denotes the corresponding margin-level ground truth tumor cellular density class, chosen from a total of $C$=7 histopathological gradings. Although the ground truth labels are defined at the margin level, FLIm measurements are acquired at the point level, resulting in a total of $n = \sum_{i=1}^{N} n_i$ individual FLIm points. Each point $x_{i,j} \in \mathbb{R}^{38}$ denotes the 38-dimensional feature vector FLIm point. For all points within a given margin $i$, the margin-level label $Y_i$ is assigned as the point-level label $y_{i,j}$, such that:

$$y_{i,j} = Y_i, \forall j \in \{1, \dots, n_i\} \quad (1)$$

To identify the potential low-confidence FLIm points, CL approach was applied on the posterior probabilities of the base model. Because these posterior probabilities directly influence the construction of the confident joint matrix, CL's effectiveness depends on the reliability of the base model. To ensure robustness, we first performed model-centric selection to identify a high-performing baseline model. Given the class imbalance in the diffuse glioma FLIm dataset, particularly the overrepresentation of margins with low tumor cellular density (see Table 2), the model selection process prioritized classifiers with balanced performance across all classes. Multiple classifiers were evaluated, and the model with the highest overall balanced accuracy and consistently high area under the receiver operating characteristic curve (AUC) across all classes was selected as the baseline. This selection was formalized as:

$$M^* = arg \max_{M \in \mathcal{M}} [\frac{1}{2}(Accuracy(M) + \frac{1}{C}\sum_{c=1}^{C} AUC_c(M) \quad (2)$$

where $\mathcal{M}$ is the set of candidate models, $AUC_c(M)$ is the AUC for class $c$, $Accuracy(M)$ is the overall accuracy for model $M$.

For every point $x_j$, let the posterior probabilities of the base model be:

$$\hat{p}_j = [\hat{p}_{j,1}, \hat{p}_{j,2}, \dots, \hat{p}_{j,C}] \quad (3)$$

where $\hat{p}_{j,c} = P(y = c \mid x_j)$ is the predicted probability that $x_j$ belongs to class $c$, and the probabilities sum to 1:

$$\sum_{c=1}^{C} \hat{p}_{j,c} = 1 \quad (4)$$

To quantify model confidence, a confidence score (CS) was computed for each point as the highest predicted class probability:

$$CS_j = \max_{1 \le c \le C} \hat{p}_{j,c} \quad (5)$$

For direct comparison with the histopathological margin label, the point-level confidence scores were aggregated into a margin-level margin confidence score (MCS), defined as:

$$MCS_i = \frac{1}{n_i} \sum_{j=1}^{n_i} CS_j \quad (6)$$

To detect mismatch between predicted and observed labels, we constructed a confident joint matrix $\hat{C} \in \mathbb{R}^{CXC}$. Each element $\hat{C}_{i,j}$ represents the number of FLIm points with observed label $y_k = i$ that were confidently predicted as class $j$:

$$\hat{C}_{i,j} = \sum_{k=1}^{n} \mathbb{I}(y_k = i).\,\mathbb{I}(\hat{y}_k = j).\,\mathbb{I}\big(CS_k \geq \tau_j\big) \qquad (7)$$

Here, predicted label $\hat{y}_k = \arg max_{c \in y}\, \hat{p}_{k,c}$, $\tau_j$ is a class-specific confidence threshold computed as the average predicted probability for class $j$, and $\mathbb{I}(\cdot)$ denotes the indicator function.

While CL is commonly used to assess label quality with respect to the assigned ground truth [21], [22], beyond potential histopathological mislabeling, confidence score in this complex clinical dataset may be impacted by other factors that affect FLIm signal as described earlier. To identify such instances, we flagged FLIm points falling into the off-diagonal entries of the confident joint matrix as high-confidence disagreements and interpreted as low-confidence FLIm points (LC):

$$LC_k = \mathbb{I}(\tilde{y}_k \neq \hat{y}_k) \cdot \mathbb{I}(CS_k \geq \tau_j) \qquad (8)$$

To characterize these low-confidence FLIm points, SHAP (SHapley Additive exPlanations) analysis was conducted to identify the most influential FLIm features driving classification decisions for each class. Based on the insights from this targeted analysis as well as patterns of class confusion observed in the classification results, an iterative class refinement process was implemented to merge adjacent classes for improved performance. After the final refinement step, $LC_k$ FLIm points were pruned from the dataset. Within the remaining dataset, class labels were regrouped where appropriate. The model was retrained on this curated high-fidelity dataset and cross-validated against histopathological ground truth to confirm biological relevance and mitigate the risk of overfitting.

To investigate potential histopathological mislabeling, margins were categorized based on the proportion of low-confidence points per margin as shown in (9). A margin was flagged as having a label issue if more than 70% of its FLIm points were labeled low-confidence using the confident joint criteria. Conversely, margins with fewer than 30% low-confidence points were designated as controls:

$$Label\ Issue_i = \begin{cases} 1, if\ \dfrac{1}{n_i}\displaystyle\sum_{j=1}^{n_i} LC_j > 0.70 \\ 0, if\ \dfrac{1}{n_i}\displaystyle\sum_{j=1}^{n_i} LC_j < 0.30 \end{cases} \qquad (9)$$

### E. Experimental Design and Evaluation

To evaluate the performance of the baseline multi-class classifier, five candidate models (random forest (RF), light gradient boosting machine (LGBM), XGBoost (XGB), multilayer perceptron (MLP), and support vector machine (SVM)) were trained and compared using a leave-one-patient-out (LOPO-CV) strategy, ensuring that data from each patient was held out during training to prevent data leakage and overfitting. For each model, we computed the overall accuracy and class-wise AUC. The model achieving the highest combination of accuracy and AUC was selected as the baseline classifier for downstream CL-based FLIm data assessment and refinement.

To investigate the low-confidence FLIm data and improve classification robustness, we evaluated the confusion matrix, overall accuracy, confidence scores, and percentage of $LC$FLIm points at each refinement step. Adjacent classes exhibiting the highest misclassification and lowest confidence (CS) were iteratively merged, and the model retrained at each step until no further improvement in performance was observed.

Finally, the label issues and control margins chosen for re-evaluation were re-submitted to the same neuropathologist for re-scoring of tumor cellular density, independent of the original histopathological assessments and blinded to prior FLIm predictions. The re-scored labels were then compared with the original FLIm-based predictions to assess whether the flagged margins indeed reflected underlying labeling inconsistencies.

## III. Results

TABLE 3
Baseline Model Evaluation

| Model | Accuracy (%) |
|---|---|
| Random Forest (RF) | 25.45 |
| Light Gradient Boosting Machine (LGBM) | 24.78 |
| Extreme Gradient Boosting (XGBoost) | 22.70 |
| Multilayer Perceptron (MLP) | 43.92 |
| Support Vector Machine (SVM) | 27.47 |

The baseline model was selected through model-centric development on the 7-class dataset evaluating five commonly used classifiers. The performance of five baseline classification models was assessed in terms of classification accuracy (Table 3). Baseline classification accuracy ranged from 22.7% to 43.9%, with the MLP model achieving the highest performance. MLP was established as the baseline multi-class classification model for downstream analysis.

### A. CL-based Class Refinements

To identify and maximize the discriminative power of FLIm in distinguishing adjacent tumor cellular density classes, we implemented a stepwise class refinement process by merging adjacent classes with high confusion and low confidence scores. The results are shown in Fig. 2. In the original 7-class configuration, substantial misclassification was observed between ‘absent’, ‘very low’, and ‘low’ tumor cellular density classes, which also exhibited the lowest confidence scores (mean CS: 0.57, 0.57, and 0.52, respectively). These three classes were merged into a single ‘low’ class to form a 5-class scheme. Training the model with 5 classes resulted in a 32% improvement in overall accuracy and boosted the merged class’s mean CS to 0.84.

Further class merging was performed to consolidate ‘low’ and ‘low-moderate’ into one class, and ‘moderate-high’ and ‘high’ into another, yielding a 3-class model. This merging further improved the class separability and overall performance: classification accuracy rose to ~83%, with class-wise accuracy of 86.0% for ‘low’, 65.8% for ‘moderate’, and 69.5% for ‘high’. Corresponding mean CS also improved (low = 0.90, moderate = 0.79, high = 0.84). The distribution of FLIm points in this final model shifted markedly toward higher confidence, indicating improved classifier calibration and reduced label ambiguity. The bar chart comparison confirmed that the 3-class scheme had the lowest percentage of low-confidence FLIm

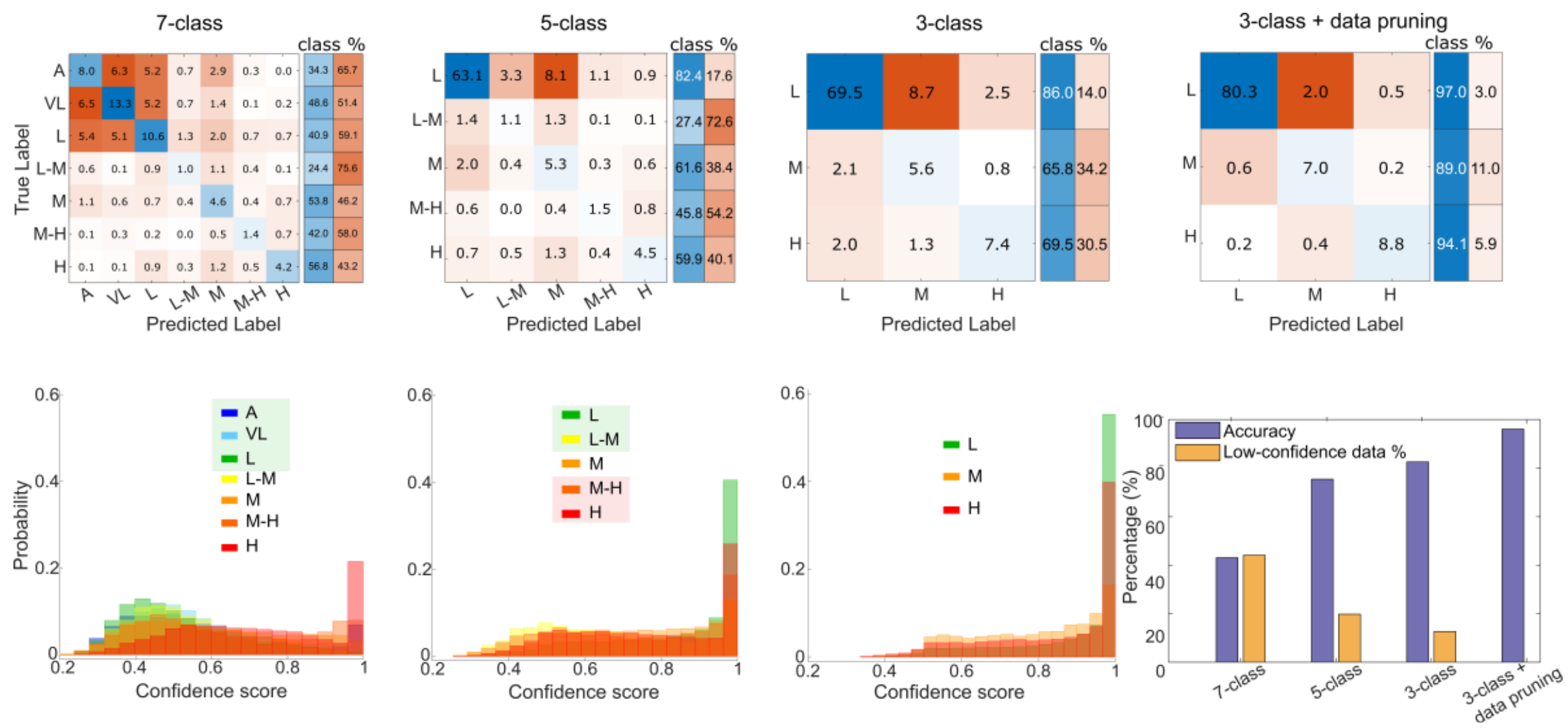

Fig. 2: Classification performance and data quality analysis for FLIm-based tumor cellular density prediction across three class granularities. Top row: Confusion matrices for the original 7-class scheme, the merged 5-class and 3-class schemes, and the 3-class scheme with low-confidence data pruning applied. Each matrix shows global accuracy percentages (cells) and per-class accuracy percentages (side bars). Bottom row: Histograms of point-level confidence scores for each class scheme, with colors corresponding to true classes. Bottom right: Comparison of overall classification accuracy (blue bars) and proportion of low-confidence data points (CS < threshold, orange bars) for each scheme. A – absent, VL – very low, L – low, L-M – low-moderate, M – moderate, M-H – moderate-high, H – high.

points (~13%) and the highest accuracy (~83%) among the three schemes.

Similar results were achieved when the original 7-class labels were regrouped into 5-class and 3-class schemes without retraining the model, suggesting that irrespective of the modeling, the FLIm data shows similar accuracies to tumor cellular densities.

To further enhance reliability, we removed the remaining low-confidence FLIm points (~13% of the data) to generate a high-fidelity 3-class training set. This pruning step eliminated noisy or uncertain samples identified via CL, particularly at the boundaries of adjacent classes. Retraining the classifier with this curated training dataset led to a substantial performance gain, boosting classification accuracy from ~83% to ~96% when tested on the same test data, demonstrating that filtering out low-confidence data points from the training set can significantly enhance model precision. This underscores the value of a DC-AI approach: not only does class merging improve model efficacy, but selective data refinement ensures that the classifier learns from the most biologically consistent and confidently labeled examples. Together, these steps resulted in a robust FLIm-based model capable of more accurate intraoperative tumor margin assessment.

## B. FLIm Feature Importance Analysis

Fig. 3 presents class-specific SHAP summary plots for the 3-class FLIm classifier, with panels (left), (center), and (right) corresponding to the low, moderate, and high tumor cellular density classes, respectively. On the X-axis, SHAP values indicate the impact of each feature on the log-odds of the given class (negative values decrease, positive values increase probability), while feature values are color-coded from low (blue) to high (pink). On the Y-axis, the most significant FLIm features are ranked from top to bottom.

SHAP analysis revealed distinct FLIm features driving the classification of tumor cellular density across classes. In the ‘low’ and ‘moderate’ classes, the Laguerre coefficient $LC2_{470\text{–}nm}$ FLIm feature consistently exhibited the highest contribution to model predictions. Although the direction of influence differed (positive for low and negative for moderate), the prominence of this feature indicates that $LC2_{470\text{–}nm}$ plays a central role in discriminating between low and moderate tumor

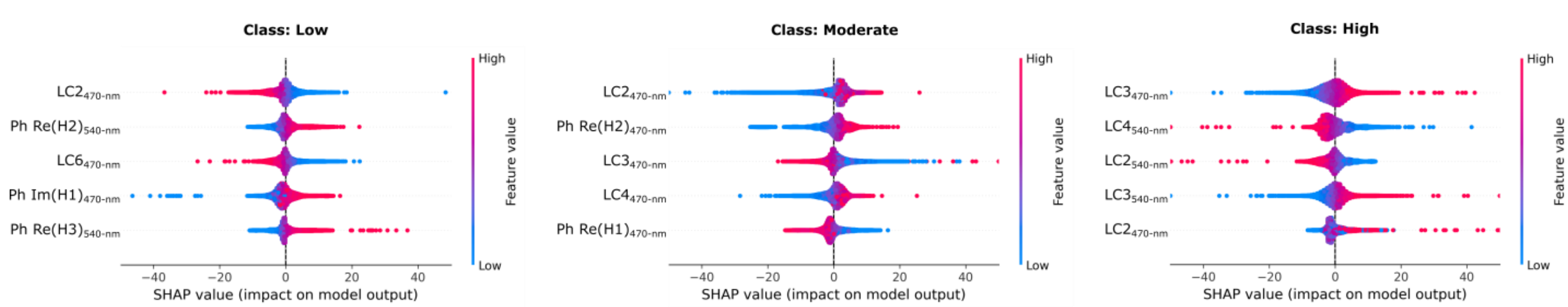

Fig. 3: SHAP summary plot illustrating the overall feature contributions of the 3-class model. Each violin represents the distribution of SHAP values for one feature, ranked by importance from top (most significant) to bottom (least significant). A positive SHAP value indicates a feature’s contribution toward a positive-class classification, whereas a negative value indicates a contribution toward a negative-class classification. The color scale (blue to pink) denotes low to high original feature values, respectively.
LT – Average lifetime | IR – intensity ratio | LC –Laguerre coefficient | Ph - Phasor

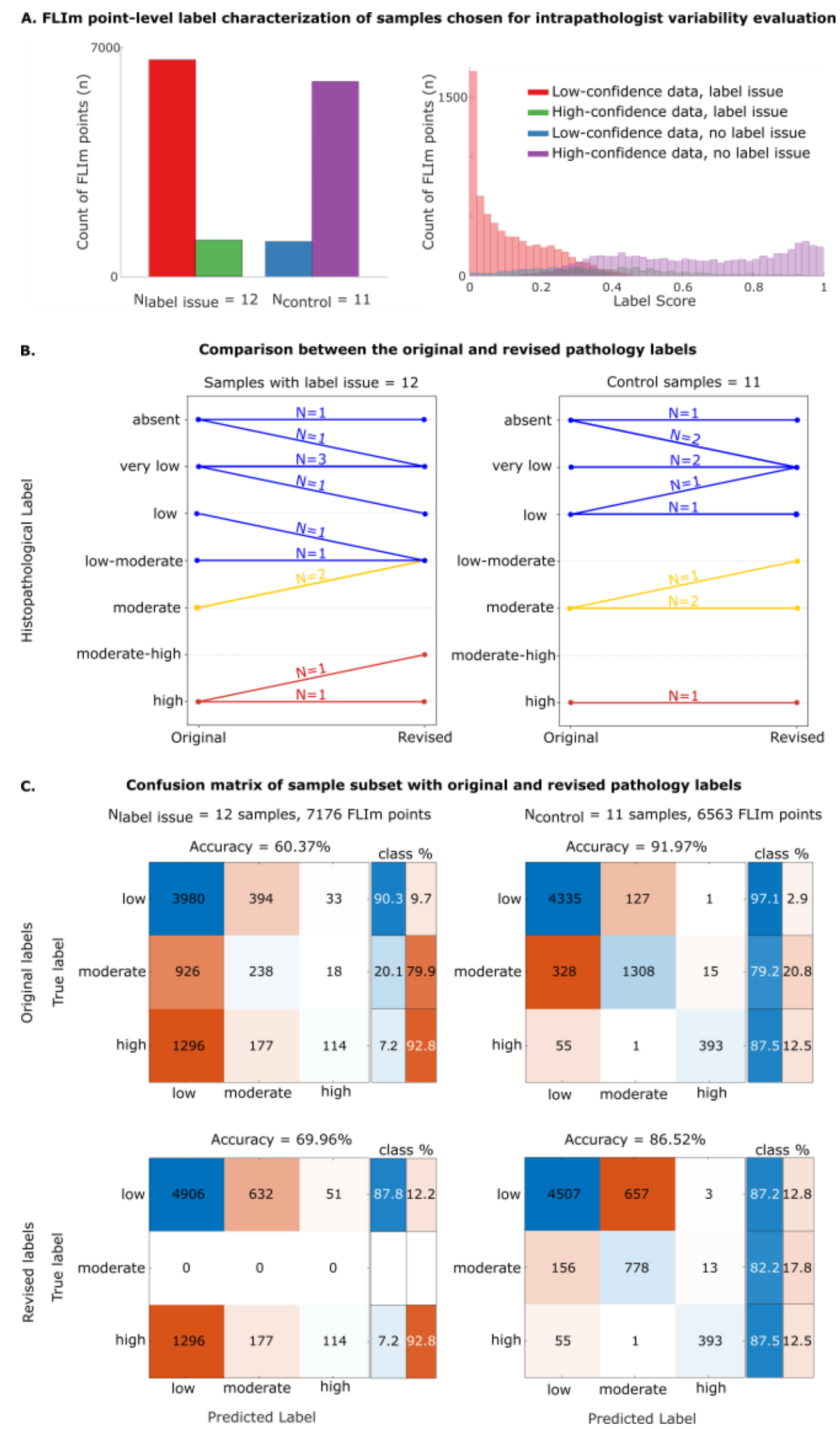

Fig. 4: Intra-pathologist variability evaluation. (A) Histograms showing the number of low-confidence and high-confidence FLIm points in margins flagged for label issues (N = 12) versus control margins (N = 11). (B) Margin-level label changes after blinded re-evaluation by the neuropathologist, shown as “spaghetti” plots of original versus revised cellular density classes for the label-issue group (left) and control group (right). Panel (C) presents confusion matrices for the 3-class classifier before (top) and after (bottom) re-scoring in the label-issue group.

cellular densities.

In contrast, for the ‘high’ tumor cellular density class, the Laguerre coefficient $LC3_{470\text{–}nm}$ emerged as the most influential feature. This shift in dominant predictor from $LC2_{470\text{–}nm}$ to $LC3_{470\text{–}nm}$ suggests that distinct decay-related parameters contribute to the separation of high-cellular density regions. Together, these findings indicate that the model relies on different subsets of FLIm features across the cellular density spectrum, reflecting the modality’s sensitivity to subtle signal variations associated with increasing tumor infiltration.

### C. Intra-pathologist Variability in Label Refinements

Intra-pathologist variability was evaluated by comparing margins flagged for potential label issues (those with >70% low-confidence FLIm points, N = 12) against control margins (<30% low-confidence points, N = 11). As shown in the bar chart (Fig. 4A, left), 85.9% of the points in the label-issue group were low-confidence (6161 out of 7176), compared to only 13.7% in the control group (897 out of 6563). The corresponding confidence score histograms (Fig. 4A, right) reveal that low-confidence points in the label-issue group cluster near zero, while high-quality control points concentrate near one. Following blind re-evaluation by the same neuropathologist, 6 of the 12 label-issue margins were assigned revised tumor cellular density labels (Fig. 4B, left), and 4 of 11 in the control set (Fig. 4B, right). In the label-issue group, this targeted relabeling improved three-class accuracy from 60.37% to 69.96% (Fig. 4C, left), demonstrating that correcting true label inconsistencies can meaningfully enhance model performance, even though overall accuracy remained modest due to continued heterogeneity. By contrast, re-scoring the control margins where labels were already reliable reduced accuracy from 91.97% to 86.52% (Fig. 4C, right), indicating that unnecessary relabeling can introduce discordance when initial annotations are sound. Together, these findings validate that CL-based flagging effectively identifies margins that benefit from neuropathologist review, while avoiding counterproductive re-evaluation of well-labeled data.

## IV. Discussion

This study leveraged a DC-AI framework to refine multi-class FLIm-based glioma classification through class merging, feature interpretation, and targeted label re-evaluation. Although contemporary deep learning architectures have demonstrated strong performance in image and waveform-based biomedical applications, the FLIm representation used here is in the form of tabular data consisting of optical features derived from Laguerre-based decay parameterization and phasor-domain analysis rather than high-dimensional raw waveforms. Prior studies have shown that deep neural networks underperform on tabular datasets compared to classical tree ensemble models [23], [24], [25]. Moreover, prior work in clinical head and neck FLIm datasets reported that a random forest classifier trained on FLIm decay features outperformed a 1D-CNN applied directly to raw fluorescence waveforms [26], further supporting the use of feature-based representations. Within this context, our objective was not to benchmark increasingly complex architectures, but to systematically investigate how a data-centric refinement under imperfect histopathologic supervision improves classification reliability. While alternative noise-robust paradigms such as uncertainty-aware learning and probabilistic label modeling could also be explored, a systematic comparison of these approaches was beyond the scope of the present study and will be investigated in future work.

This methodological framework is not limited to FLIm; rather, it illustrates principles that may also benefit other intraoperative optical platforms such as Raman spectroscopy and optical coherence tomography [27], [28], where data variability and label uncertainty similarly challenge model development. The following sections discuss establishing a baseline classification model, key sources of low confidence in FLIm predictions and how targeted pathologist review can help address them.

## A. Baseline Model Performance Evaluation

The CL framework depends on a well-calibrated baseline classifier to identify potentially mislabeled or low-confidence data points. However, most clinical datasets exhibit significant class imbalance, with data distribution heavily skewed towards low tumor cellular density. If the baseline model is poorly chosen or inadequately trained, it may become biased toward the majority class, thereby undermining the reliability of the confident joint. Consequently, rigorous model-centric development ensuring high predictive accuracy and robustness to class imbalance must precede data-centric operations such as pruning, relabeling, and refinement. These steps are essential for achieving meaningful data curation and downstream interpretability.

We compared five widely used complementary modeling approaches, including linear (SVM), decision-tree (RF), boosting methods (LGBM, XGB), and neural networks (MLP) [29], to identify strategies best suited for FLIm data. Among these, the MLP achieved more than 20% improvement in classification accuracy relative to the other models (Table 3). These results underscore the importance of a well-calibrated baseline model as a prerequisite for reliable confident learning, ensuring that subsequent data-centric refinements are anchored in a robust and minimally biased modeling foundation.

## B. Targeted FLIm data analysis

### i. Effect of Brain Tissue Type

To investigate the factors contributing to low-confidence FLIm predictions, we conducted a detailed analysis of a representative patient from the 3-class model, in whom 43.3% of FLIm points were flagged as low-confidence scores. All margin samples for this patient were labeled as having ‘low’ tumor cellular density. Further analysis revealed a clear dependance of classification confidence on tissue composition, specifically whether the sampled region consisted of grey matter, white matter, or a mixture of both. FLIm points originating from grey matter consistently exhibited lower confidence scores than those from white matter (Fig. 5), a pattern that was reflected in classification performance: while white matter samples achieved high accuracy (98.2%) whereas grey matter and mixed-tissue samples exhibited substantially lower accuracy (20–25%). These differences likely reflect underlying biological and metabolic distinctions between grey and white matter. Grey matter is characterized by a higher density of neuronal cell bodies and mitochondria, which is associated with shifts in metabolic state and an increased contribution of protein-bound NAD(P)H to the fluorescence lifetime signal [30], [31]. Such tissue-specific metabolic signatures may reduce separability in regions of low cancer cellular density, particularly for models trained on datasets dominated by white matter–derived signals. In contrast, the more homogeneous microstructural and metabolic profile of white matter may yield more consistent and discriminative FLIm signatures. Collectively, these findings indicate that anatomical heterogeneity introduces biologically driven variation in FLIm signals that can adversely affect classification confidence and performance. Consistent with prior studies [17], [18], [32], [33], [34] these results highlight the importance of incorporating tissue-specific context when applying FLIm-based classification approaches in vivo.

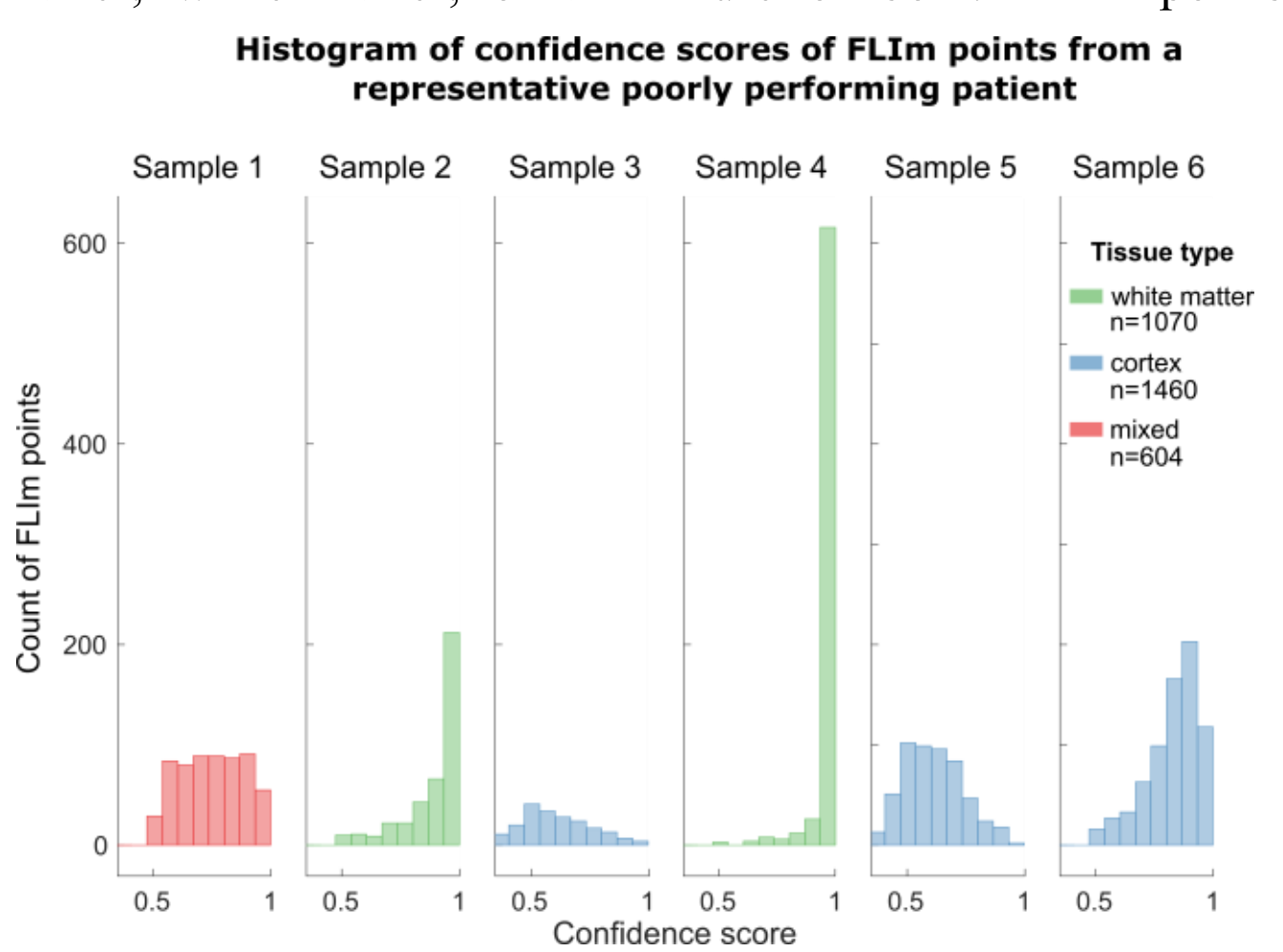

Fig. 5: Effect of brain tissue type on data quality and classification accuracy in low tumor cellular density class in a representative case: confidence scores stratified by tissue type show reduced data quality in grey matter and mixed tissues compared to white matter.
n= number of FLIm points

### ii. Effect of blood in the field-of-view

In another low-performing case from the 3-class model, 31.2% of FLIm points were flagged as low-confidence. Closer inspection of the intraoperative data revealed that two out of three collected samples had significant blood present in the field-of-view (Fig. 6). Blood absorbs and scatters light across a broad spectral range, altering the native fluorescence signal and introducing variability in the measured fluorescence decay profiles [3], [6]. This optical interference likely contributed to both the reduced confidence scores and the observed degradation in classification performance. Collectively, these observations highlight the sensitivity of FLIm signals to intraoperative imaging conditions and underscore the importance of maintaining a clear surgical field during FLIm acquisition. Moreover, they suggest that integrating real-time visual cues or signal quality-based feedback during FLIm-guided procedures may help mitigate blood-related artifacts and improve data reliability in vivo.

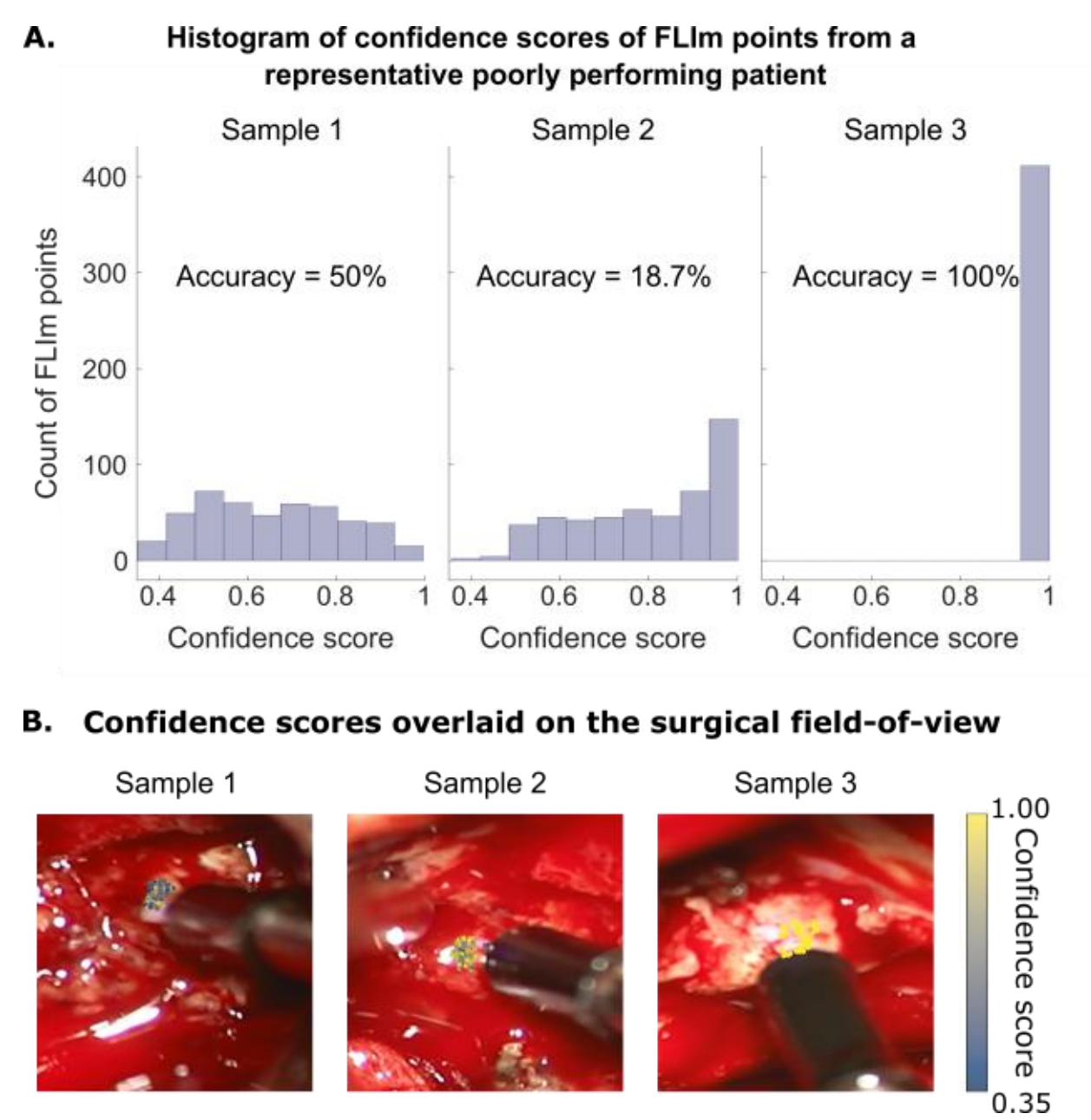

Fig. 6: Effect of blood in the surgical field-of-view on data quality and classification accuracy in a representative case.
(A) Confidence scores histogram and accuracy of the samples show reduced data quality. (B) Confidence scores overlaid on the surgical field-of-view on the samples.

### *C. Interpretation of Class-Specific FLIm Feature Contributions*

SHAP analysis identified $LC2_{470\text{–nm}}$ and $LC3_{470\text{–nm}}$ as dominant contributors to model predictions; however, their biological interpretation remains inferential. In the ‘low’ and ‘moderate’ tumor cellular density classes, $LC2_{470\text{–nm}}$ exerted the strongest influence. Given that LC2 is inversely related to fluorescence lifetime and that the 470-nm spectral band predominantly reflects NAD(P)H dynamics [35], the observed SHAP trends where lower $LC2_{470\text{–nm}}$ values favor 'low’ class predictions and higher values favor ‘moderate’ class predictions could be linked to shifts in the relative abundance of protein-bound versus free NAD(P)H. Specifically, healthy tissue tends to exhibit a higher proportion of protein-bound NAD(P)H with longer lifetimes, whereas tumor tissue may show an increased free NAD(P)H fraction associated with elevated glycolytic activity (the Warburg effect) [36].

In contrast, $LC3_{470\text{–nm}}$ emerged as the most influential feature in the ‘high’ tumor cellular density class, suggesting that different temporal decay components may become more discriminative with increased tumor infiltration. This shift may reflect progressive metabolic reprogramming or microenvironmental differences in moderate vs highly cellular tumor regions [37], [38]. Overall, these findings suggest that FLIm captures distinct fluorescence decay signatures across the tumor cellular density spectrum and may provide indirect sensitivity to underlying metabolic gradients, even though the precise biochemical basis remains to be validated through additional targeted experimental studies.

### *D. Label Quality and Re-Evaluation: A Pathologist-in-the-Loop Perspective*

The clinical significance of the revisions differed substantially between the label-issue and control groups following blinded re-evaluation by the same neuropathologist. In the control group, most revisions occurred within the lowest tumor cellularity categories, which were ultimately consolidated into a single “low” class and have limited impact on intraoperative surgical decision-making. In contrast, the label-issue group included revisions in moderate and high cellular density (tumor-positive) classes, therefore more consequential from both a clinical and modeling perspective. Although the dataset is skewed toward low-cellularity margins, resolving label ambiguity in the underrepresented moderate and high classes is particularly important, as mislabeling in these clinically significant categories can disproportionately affect classifier training and performance.

This analysis also highlights several limitations inherent to the study. First, the use of a 7-class semi-quantitative histopathological grading scheme proved unrealistic at the lower end of the spectrum (absent, very low, and low), where both biological overlap and limited FLIm sensitivity hinder reliable separation. Second, tumor cellular density grading is subject to human inconsistency, even when performed by a single expert. While programmatic cell-counting approaches could reduce variability, they are also challenged in low-cellularity settings where tumor cells are sparse and difficult to distinguish from normal tissue. Future studies incorporating multiple pathologists with inter-rater agreement metrics may help establish more robust reference standards. Third, class imbalance and the relatively small sample size used for re-evaluation (12 label-issue vs. 11 control samples) limit the magnitude of observable performance gains. Nevertheless, the primary goal of this analysis was to demonstrate a proof-of-concept data-centric framework for systematically flagging ambiguous samples, thereby reducing the burden of exhaustive manual re-annotation and focusing expert review where it is most impactful.

## V. Conclusion

This study demonstrated the utility of a DC-AI approach, incorporating confident learning methodology to identify and enhance the FLIm sensitivity for intraoperative glioma margin classification. By systematically identifying and addressing low-confidence data points through characterization, pruning, and targeted label re-evaluation, we curated a high-fidelity dataset that enhanced model robustness and biological interpretability. Future studies integrating FLIm with complementary optical modalities such as Raman spectroscopy, which provides molecular-specific contrast, can help to better contextualize FLIm-derived biochemical signals. Furthermore, our intra-pathologist analysis revealed scoring variability even within a single expert, reinforcing the need for robust ground truth frameworks in future studies. Collectively, these findings underscore the importance of integrating model-driven label refinement, data quality assessment, and feature-level interpretation to develop clinically reliable, real-time optical imaging tools for surgical guidance in glioma resection, providing a foundation that could be adapted for other intraoperative optical technologies.